# Vision-based inspection system employing computer vision & neural networks for detection of fractures in manufactured components

Sarthak J. Shetty
*Department of Mechanical Engineering*
*R.V. College of Engineering*
Bengaluru, India
sarthakjs.me15@rvce.edu.in

*Abstract*—we are proceeding towards the age of automation and robotic integration of our production lines [5]. Effective quality-control systems have to be put in place to maintain the quality of manufactured components. Among different quality-control systems, vision-based inspection systems have gained considerable amount of popularity [8] due to developments in computing power and image processing techniques. In this paper, we present a vision-based inspection system (VBI) as a quality-control system, which not only detects the presence of defects, such as in conventional VBIs, but also leverage developments in machine learning to predict the presence of surface fractures and wearing. We use OpenCV, an open source computer-vision framework, and Tensorflow, an open source machine-learning framework developed by Google Inc., to accomplish the tasks of detection and prediction of presence of surface defects such as fractures of manufactured gears.

## I. INTRODUCTION

Vision-based inspection systems (hereafter abbreviated as VBI), are employed as quality-control systems along various critical points in the production line [8], [6], [9], employing cameras for the detection of defects in manufactured components. Due to advances in processor design we now have access to vast volumes of computing power and open source libraries, utilizing robust image-processing techniques for the detection of fractures. VBIs can analyze various features of the manufactured component through a camera feed interfaced with a computer system for the analysis of the same. Irregularities are highlighted by the system and are visible on the display paired with the system. The image is then passed onto a fracture-prediction system which employs a machine learning framework and provides a probabilistic value of whether or not the given component is fractured or not.

This combination of image-processing techniques and machine learning tools serves dual processes: a) The OpenCV based computer-vision sub-system provides a visual reference for the human operator allowing them to check, and counter-check for defects which may or may not be apparent to the machine-learning sub-system. b) Inception V3, a Tensorflow based machine-learning model [10] with retrained final layers, provides a probabilistic value of whether a surface defect such as a fracture or defect is present or not. The machine learning model may identify defects that might not be apparent to the human operator, or are too minute for computer vision filters to effectively highlight.

## II. LITERATURE SURVEY

Edwards, M [5] has provided a review of how the manufacturing industry poses to be revolutionized by the advent of robots, and also the rapid adoption of robotics for the completion of variety of tasks in production lines.

Chen, F.L. et. al [2], describes one of the first machine-vision based inspection systems along computer integrated manufacturing system (CIMS), describing it's measurement flexibility, non-destructive property and high resolution as significant advantages over other inspection systems.

Baygin, Mehmet et. al [1] has proposed a machine-vision based inspection system for identifying defects in printed circuit boards (PCBs). The system measures the dimensions of the holes drilled in the circuit board, and also checks whether the dimensions are within 2μm of the set value.

Stojanovic, Radovan et. al [9] has proposed a real-time machine-vision based inspection system which employs neural networks for the identification and classification of defects in textiles.

Malamas, N. E et. al [8] has provided a wide review on industrial machine-vision. Advantages of machine-vision based systems over human inspection have also been discussed broadly in the publication.

## III. PROPOSED APPROACH

As mentioned earlier, the VBI system that we have developed is based on two parts: a) An OpenCV based computer-vision system that applies different convolution filters on a specimen image, highlighting any fractures or defects on the surface. b) A TensorFlow based convolutional neural network model known as Inception V3 model. The final layers of this neural network are retrained on two categories of images: a) normal Gears & b) defective gears, to distinguish between normal gears and broken gears.

A Google-Image webscraper [12] was used to scrape Google Images to generate the aforementioned dataset comprising of 200 images of normal gears and 200 images of fractured gears. This dataset was then used to re-train the final layers of the Inception V3 connvolutional neural network.

A variety of filters are used for the detection of edges in images [7], [3]. In this paper, the image processing system will be employing the use of a filter-bank (also known as a kernel-bank in this context), comprising of 4 convolution-filters:

1. Sobel X
2. Sobel Y
3. Laplacian
4. Sharpen

These filters are applied on a gray-scale version of the image. Carrying out operations on grayscale images has been deemed computationally lightweight for embedded systems such as FPGAs, which are often employed for machine-vision tasks in production lines [8].

Once the image has been converted into grayscale, it is subjected to the above filters. The filters are convolution matrices which, through simple convolution operations with the grayscale images generate an output image in which the edges of the component are highlighted. Since the matrices have the ability to highlight edges, they can highlight the fractures which are present on the surface of the component.

The image is then passed on to the prediction model, which provides a probabilistic prediction of a fracture or crack being present on the surface.

## IV. OVERVIEW OF MODEL

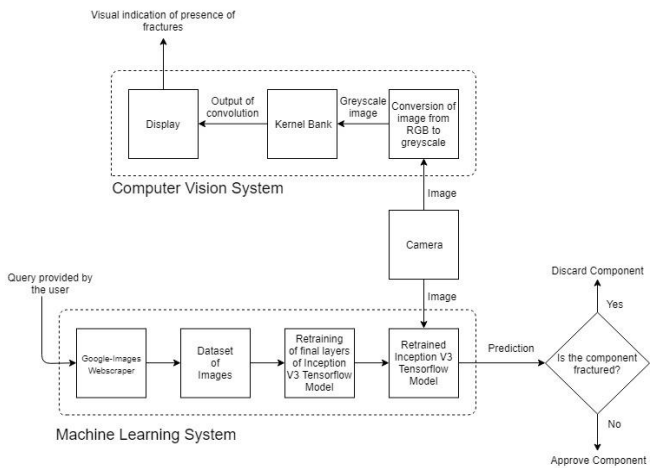

Fig. 1. Overview of model

## V. CONVERSION FROM RGB TO GREYSCALE

Initially, the image is converted to gray-scale from RGB, since it is computationally more light-weight and faster to process. For example, below we have provided two sets of images to an example filter; the image on the left is obtained by applying the "Sobel X" convolution filter on a grayscale image of the gear, whereas the image on the right is obtained by applying the "Sobel X" convolution-filter on a color image of the gear.

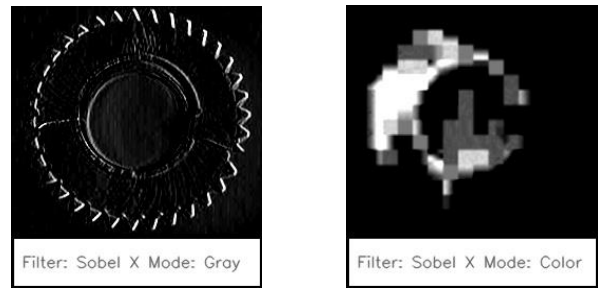

Fig. 2. Comparision of convolution output for a grayscale image (L) input versus output for a color image input (R)

As we can notice, the features of the grayscale image are recognizable, whereas the features in the color image are indecipherable. The image obtained from this process is used by the remaining stages of the computer vision sub-system. However, since the machine-learning model was trained on color images, the color image of the gears is passed onto the machine-learning subsystem.

## VI. APPLYING GAUSSIAN BLUR

The color image of the component (obtained through a camera feed above the production line), is likely to have a lot of noise associated with. Before proceeding with the rest of steps, we apply a Gaussian blur on the image, smoothen it, and thereby reducing the associated noise.

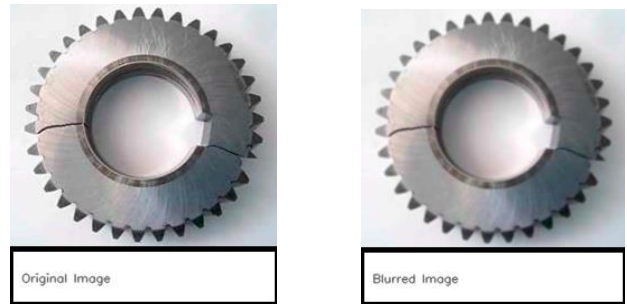

Fig. 3. Comparision of original [L] and resultant image after applying Gaussian blur (R)

However, the stand deviation of the Gaussian blur applied on the image should not be increased incessantly. In some cases blurring the image too much leads to smudging of the edges, once the filters have been applied. [Fig. 4].

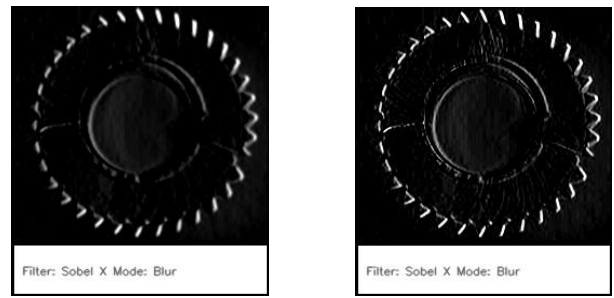

Fig. 4. Comparison of Sobel-X filtered images, when the standard deviation of blurring (along X and Y axes) is 13 (L) and 3 (R)

## VII. KERNEL-BANK

The convolution-filters mentioned earlier are also referred to as "kernels". These kernels are two dimensional matrices, which, through a series of mathematical operations with the grayscale image, produce output images with highlighted fractures. In this system, we employ 4 different kernels, which are:

### A. Sobel-X kernel

This kernel is employed for detection of horizontal edges in images. It is a 3x3 matri which is convolved with the grayscale image to obtain a resultant image in which the horizontal edges are highlighted. It is represented as '$G_x$'.

$$G_x = \begin{bmatrix} -1 & 0 & +1 \\ -2 & 0 & +2 \\ -1 & 0 & +1 \end{bmatrix} * I$$

Fig. 5. Matrix representation of Sobel-X operator *(Note: 'I' is the matrix representation of grayscale image)*

### B. Sobel-Y kernel

Similar to the Sobel-X operator, the Sobel-Y operator is also used for edge detection. It is also a 3x3 matrix, which when convolved with the grayscale image to obtain a resultant in which the vertical edges are highlighted. It is represented as '$G_y$'.

$$G_y = \begin{bmatrix} -1 & -2 & -1 \\ 0 & 0 & 0 \\ +1 & +2 & +1 \end{bmatrix} * I$$

Fig. 6. Matrix representation of Sobel-Y operator *(Note: 'I' is the matrix representation of grayscale image)*

### C. Laplacian Kernel

The Laplace operator obtains the second derivative of the pixel intensity. Once the Sobel operators are applied on an image, the gradient of the pixels at the edge are the highest. Hence, the second derivative of the intensity at this point will be 0, and hence the edge can be detected.

$$\text{Laplace}(f) = \frac{\partial^2 f}{\partial x^2} + \frac{\partial^2 f}{\partial y^2}$$

Fig. 7. Laplace equation for determining second derivative of pixel intensity

### D. Sharpen Kernel

Images are usually composed of fine details, and these details tend to have high frequency components that make the image richer. If these high frequency components are attenuated, details of the images tend to be lost. Hence, in order to exaggerate the details and edges of the images, we pass the grayscale image through a sharpening kernel.

## VIII. RESULTS OF CONVOLUTION

TABLE I. IMAGES OBTAINED AFTER CONVOLUTION

| Serial No. | Images obtained after convolution | |
|---|---|---|
| | *Kernel Applied* | *Result of convolution* |
| 1. | Sharpen | 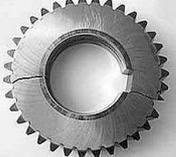 |
| 2. | Sobel X | 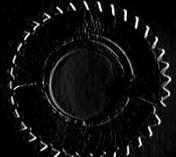 |
| 3. | Sobel Y | 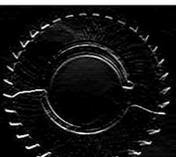 |
| 4. | Laplacian | 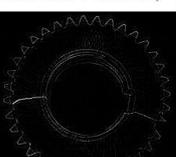 |

## IX. GENERATING DATASET FROM GOOGLE IMAGES

In order to train the TensorFlow model, a large dataset of images has to be provided. In this paper, we wish to distinguish between broken and normal gears. Hence, we use a Google Images webscraper [12] to generate the two datasets of 200 images each; a) an image dataset of normal gears b) an image dataset of broken gears.

## X. RETRAINING FINAL LAYERS OF INCEPTION V3

Inception V3 is a TensorFlow based machine-learning model, trained on a dataset of images belonging to 1000 classes, with an identification error rate of just 3.4% (for comparison, humans exhibit an identification rate of 5.1% on the same dataset) [11].

In order to prevent the model from memorizing the features in the dataset ("overfitting") [4], the dataset is split into a) training, b) validation & c) test set. The training of the model is accordingly judged using three metrics; a) training accuracy, b) validation accuracy & c) cross entropy.

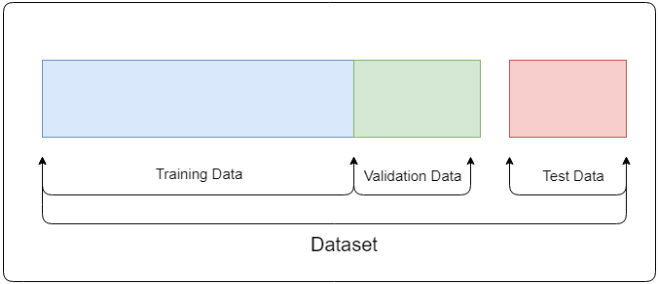

Fig. 8. Division of datset during retraining of the model

### A. Training Data & Training Accuracy

This is the prediction accuracy that the model attains when it's being trained on the portion of the dataset that is reserved as the training set. It comprises of nearly 60% of the image dataset, on which the model is trained.

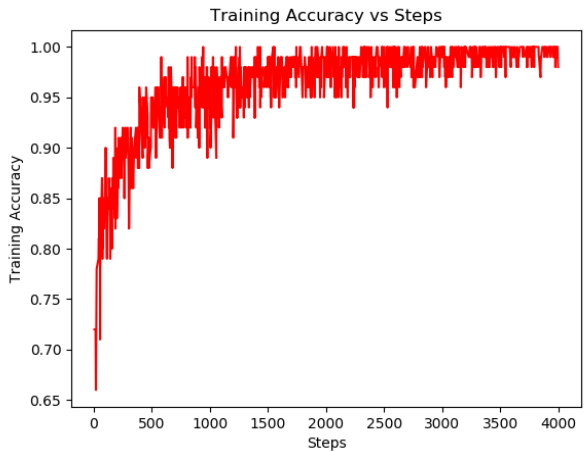

Fig. 9. Increase in training accuracy with number of steps during retraining

### B. Validation Data & Validation Accuracy

The validation data is used to keep the model in check during training, to make sure that it is not overfitting on the data provided, and randomly views it during training.

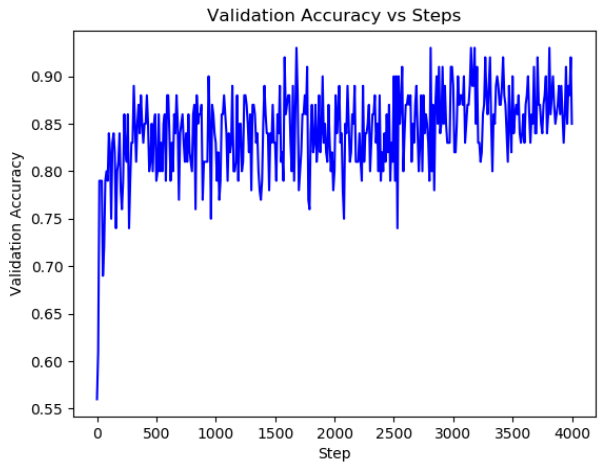

Fig. 10. Increase in validation accuracy with number of steps during retraining

Accordingly, the validation accuracy is the measure of how well the model correctly labels the images when the model is being trained on the validation data.

### C. Test Data

This portion of the data is kept away from the training data & validation data during the training period. This data is used to determine the final accuracy of the model, once the training process on the other two sets has completed. Hence, the model will be making predictions on previously unseen data.

### D. Cross-Entropy

This metric is used to determine how close the distribution of predictions made by the model is to the true distribution. Our aim is to reduce the cross-entropy as much as possible (as shown in Fig. 11.), to ensure that the predictions made by the model closely resemble the actual labels in the dataset.

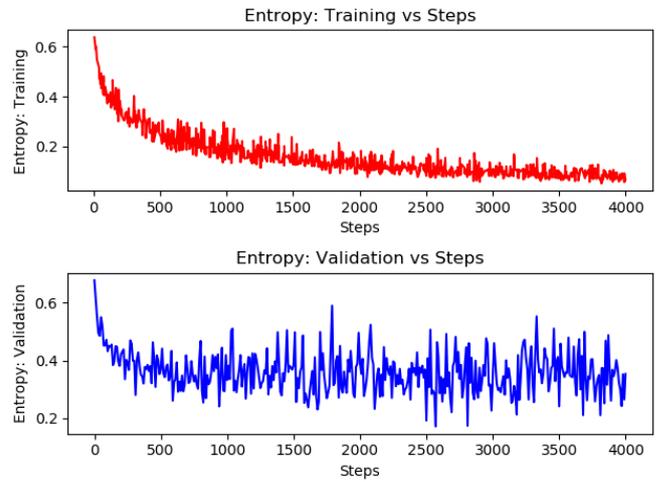

Fig. 11. Gradual decrease in cross-entropy during training and validation with number of steps during retraining.

## XI. PREDICTING THE PRESENCE OF DEFECTS

When the image is routed to the machine-learning subsystem, the pre-trained TensorFlow model tries to probabilistically determine which category the gear falls into. If the probability assigned to either category (in this case, normal gear and broken gear) exceeds 0.5, then the gear is classified into that particular category.

```
[INFO]The results of the retrained model are as follows:
[INFO]Probability that the given image is a normal gear is: 0.99599946
[INFO]Probability that the given image is a broken gear is: 0.0040006186
[INFO]The given component is a: normal gear
```

Fig. 12. Result from the trained TensorFlow model

If the gear is predicted by the model to be broken, it is immediately discarded from the production line. In some cases however, the system can wrongly predict the gear to be fractured; in such a case, the computer-vision system can be used by a human operator to visually check for the presence of defects.

## XII. RESULTS OF RETRAINED MODEL

In most the cases the machine-learning sub-system correctly predicts the label of the component; as shown below in Table II. However, in some cases a label may be wrongly assigned to the given component. This feature of the system has been highlighted in the table below as well.

TABLE II. RESULTS FROM SYSTEM

| Serial No. | Prediction obtained from VBI | |
| --- | --- | --- |
| | *Input Image* | *Predictions* |
| 1. | 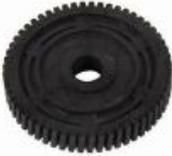 | [INFO]The results of the retrained model are as follows:<br>[INFO]Probability that the given image is a normal gear is: 0.91759664<br>[INFO]Probability that the given image is a broken gear is: 0.08240334<br>[INFO]The given component is a: normal gear |
| 2. | 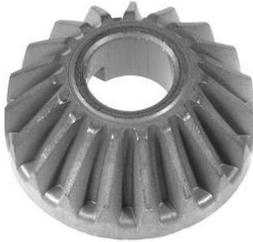 | [INFO]The results of the retrained model are as follows:<br>[INFO]Probability that the given image is a broken gear is: 0.86653554<br>[INFO]Probability that the given image is a normal gear is: 0.1334645<br>[INFO]The given component is a: broken gear |
| 3. | 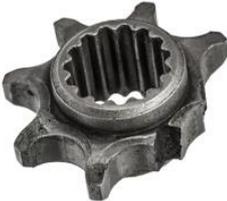 | [INFO]The results of the retrained model are as follows:<br>[INFO]Probability that the given image is a broken gear is: 0.969334<br>[INFO]Probability that the given image is a normal gear is: 0.030665962<br>[INFO]The given component is a: broken gear |
| 4. | 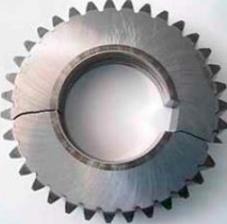 | [INFO]The results of the retrained model are as follows:<br>[INFO]Probability that the given image is a broken gear is: 0.83579403<br>[INFO]Probability that the given image is a normal gear is: 0.16420601<br>[INFO]The given component is a: broken gear |

a. Images shown were not part of the initial dataset that the model was trained on

For instance, in the second example, the model has incorrectly labelled the gear as broken, even though there are no apparent signs of wear or defects. Apart from this, the model does fairly well, clearly distinguishing between normal and broken gears.

## XIII. SCOPE FOR FURTHER DEVELOPMENT

As seen in the previous section, the model can incorrectly label a given component (Table II, serial no. 2). This is primarily due to the size of our dataset used to retrain the model. In order to improve the predictions generated by this model, we have to increase the size of this dataset and introduce the model to more and more examples, thereby increasing the variety of samples that the model has previously encountered. In order to further improve the detection of fractures by the computer-vision subsystem, we can incorporate a wider array of filters, to reduce the noise without necessarily reducing the quality and the richness of the image in the process.